\documentclass[11pt,a4paper]{article}
\usepackage[hyperref]{acl2019}
\usepackage{times}
\usepackage{latexsym}
\usepackage{amsmath}
\usepackage{amssymb}
\usepackage{graphicx}
\usepackage{subfig}
\usepackage[normalem]{ulem}

\usepackage{comment}
\usepackage{color}

\usepackage{url}

\aclfinalcopy 

\title{Can Neural Image Captioning be Controlled via Forced Attention?}

\author{Philipp Sadler, Tatjana Scheffler and David Schlangen \\
	Department of Linguistics\\
	Research Focus Cognitive Sciences\\
	University of Potsdam, Germany \\
	\texttt{\{name.surname\}@uni-potsdam.de}
}
\date{}

\begin{document}
\maketitle
\begin{abstract}
	Learned dynamic weighting of the conditioning signal (attention) has 
	been shown to improve neural language generation in a variety of 
	settings. The weights applied when generating a particular output 
	sequence have also been viewed as providing a potentially explanatory 
	insight into the internal workings of the generator. In this paper, we 
	reverse the direction of this connection and ask whether through the 
	control of the attention of the model we can control its output. 
	Specifically, we take a standard neural image captioning model that uses 
	attention, and fix the attention to pre-determined areas in the image. 
	We evaluate whether the resulting output is more likely to mention the 
	class of the object in that area than the normally generated caption. 
	We introduce three effective methods to control the attention and find
	that these are producing expected results in up to 28.56\% of the cases. 
\end{abstract}

\section{Introduction}

Sequential deep learning language models with an attention mechanism are able to use not only the immediately previous inputs, but can involve context by ``attending to''  select parts of the whole input sequence at each time step. This has first been shown to be helpful for neural machine translation, which operates on sequences of words. Here, deep learning networks with attention are capable to jointly learn the alignment and translation of languages \cite{bahdanau_neural_2014, luong_effective_2015}.

\begin{figure}
	\centering
	\vspace{5mm}
	\includegraphics[width=0.45\textwidth]{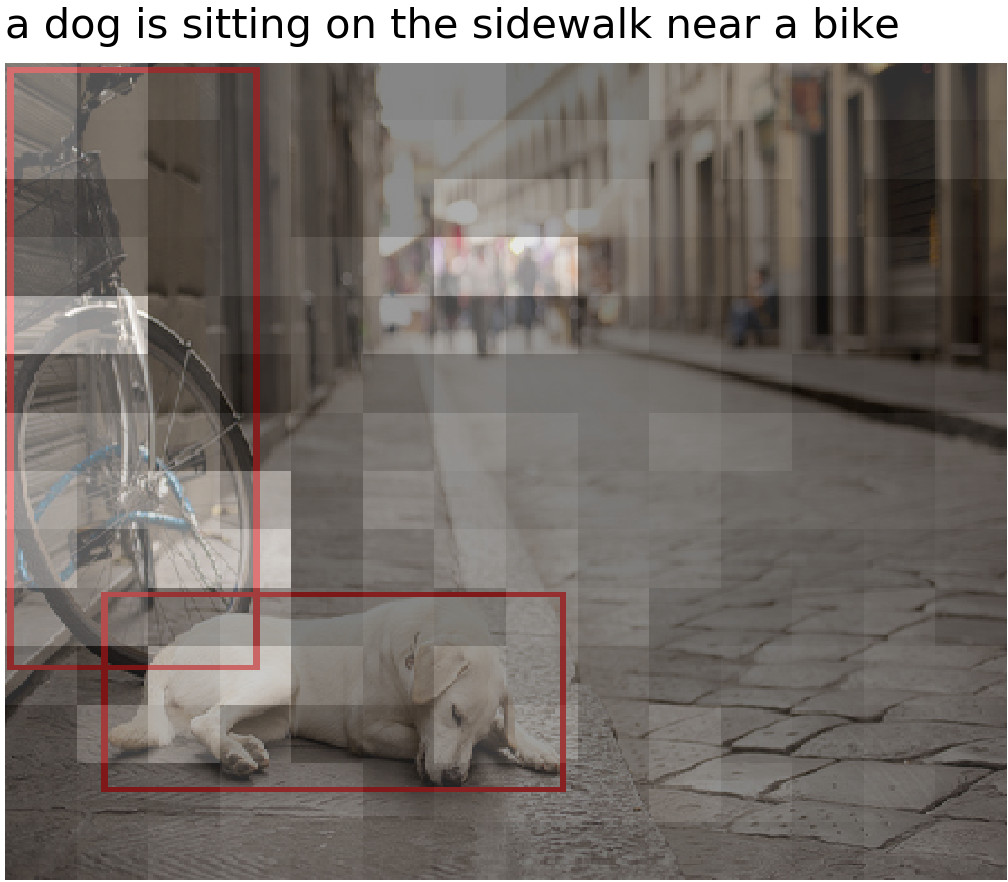}
	\caption{A caption generated by an image captioning model. The attention is pixelated and summed up over all time steps. In addition, the dog and the bicycle are framed with the corresponding bounding boxes.}
	\label{fig:dog_boxes}
\end{figure}

From the beginning, the dynamics of the attention while generating output sequences has been seen as providing insight into the workings of the models, if only qualitatively. This is in particular applicable to the  interdisciplinary fields of natural language processing and computer vision like image captioning. For example, \citet{xu_show_2015} overlay the attention weights over the input image and show how it shifts while generating the image caption step by step. We show a variant of this visualisation type in Figure~\ref{fig:dog_boxes}. The implicit argument at least seems to be that this is informative, because there is a causal relation between where the attention is placed and what is being produced, which has been critically discussed recently \cite{serrano_is_2019,jain_attention_2019}.

In this paper, we address the question of whether this assumed connection can be used to assert additional control over the generation process. We train a caption generation model with spatial attention in the usual way, but then at test time override its attention mechanism and force it to attend to pre-determined parts of the image. Does this cause the generated output to be different from what would otherwise have been produced, in predictable ways? Our results show that this expectation is partially supported. 


\section{Related Work}

There have been several attempts to achieve more control over neural language generation. \citet{anderson_guided_2017} control the output of a captioning model at test time with an enhanced beam search. An external system is generating image tags as a control signal at the decoder level. They show that adding the additional hints for the generation process actually improves the performance for out-of-domain captioning. Although this approach works, there is no attention effecting mechanism involved. \citet{zarriess:inlg18} evaluate a ``trainable decoding'' approach that inserts task-specific concerns into the decoding process.

Interfering with the attention mechanism after training, however, has to our knowledge been tried less often. \citet{cornia_show_2018} train a captioning model not only to learn the distribution for images and sentences, but also for bounding boxes and noun chunks. In addition, the model has to learn when to switch between boxes. As a result, the captioning model is controllable by a bounding box sequence provided as an input to the network at test time. Although this approach has been shown to work well, here the model is explicitly designed to be controllable.

In contrast to the previous approaches, we assume that an attention-aware captioning model is inherently controllable by its spatial attention. In such a sense, our approach is an inverse of the visual grounding task. In one of their experiments, \citet{rohrbach_grounding_2016} try to localize phrases within an image by deriving bounding boxes from the spatial attention of a specially trained model. We try to reverse this direction and fix the attention to manually chosen parts of the image after training to generate captions about that region.

\section{An Attentive Image Captioning Model}

We implement a standard neural image captioning architecture that uses spatial attention, which has shown to correlate objects within the image with spatial attention \cite{xu_show_2015}. As a modification, we use the image features of the pooling layer in the fifth convolutional block like \citet{yang_stacked_2015}. This modification lead to higher BLEU scores for our setup. 

The network is trained on the training split of the MSCOCO dataset for the \emph{Captioning Challenge 2015}. This split provides five ground-truth captions for each of the 82,783 images. The dataset images are resized to 448x448 pixels \textit{not} keeping the aspect ratio. We apply only basic tokenization on the captions. Then the captions of the dataset are prepared to contain only captions that have a maximal length of 16 words. Furthermore, the vocabulary is constrained to the 10,000 most common words and we discard captions that are containing words not included in this vocabulary. The caption-image pairs are shuffled randomly before training. As in the work of \citet{xu_show_2015}, we apply dropout and use the Adam optimizer to minimize the loss function

\begin{equation}
L = -log(P(y|x)) + \lambda \sum_{i}^{L}{(1 - \sum_{t}^{C}{ \alpha_{it} })^2}
\end{equation}

with the alpha-regularizer $\lambda$, $L$ as the number of image features, $C$ as the caption length and $\alpha_{it}$ as the spatial attention for an image feature at a specific time-step. The alpha-regularizer $\lambda$ constraints the caption generator to distribute the spatial attention more equally among the image areas during the whole generation process. \citet{xu_show_2015} noted that this regularizer is important for the resulting overall BLEU score, but they did not mention the exact value to be chosen. Therefore we tried $0.001, 0.005, 0.010$ and found that $\lambda = 0.005$ produces the best scores.

Our best model achieves $69.8$, $51.8$, $37.2$, $26.6$ in BLEU-1,2,3,4, respectively. Thus we were able to partially increase on the reported scores on the same validation split. \citet{xu_show_2015} reported $70.7$, $49.2$, $34.4$, $24.3$, so our model is worse by $-0.9$ in BLEU-1 scores, but better by $+2.6$, $+2.8$, $+2.3$ in the other scores.

\section{Methods}

Assuming that a sufficiently well trained captioning system is capable of talking about a variety of objects and object configurations, we expect that the caption generation process is controllable when forcing the attention to specific parts in the image.  We create a dataset for this specific captioning task using the MSCOCO validation split for the \emph{Detection Challenge 2015}, which provides one or more bounding boxes for each of the 40,504 images \cite{lin_microsoft_2014}. The boxes frame distinct (but possibly overlapping) objects in the images that are each labelled with one of 80 categories.

\subsection{Constructing Spatial Attention Vectors}
\label{construct}

The distinct objects in the images are framed with bounding boxes which are rectangles defined by their width, height and the xy-coordinates of the left upper corner. We discard all bounding boxes that are smaller than the median size, because the model less likely attends to small objects in the images. From the remaining boxes we derive the spatial attention vectors for the experiments.

To construct a spatial attention vector, we define a matrix $Z \in \mathbb{Z}^{W \times H}$ with $W \times H$ as the according image size and a box coordinate space $B \in\mathbb{Z}^{2}$ with each point that falls into the box. Given these we set the values in the matrix as the following:

\begin{equation}
\alpha_{xy} = 
\begin{cases}
255 \textnormal{ if } (x,y) \in B\\
0 \textnormal{ otherwise }
\end{cases}
\label{word-use}
\end{equation}

We use $255$ as the attention values to align to RGB format, so that we can easily present the maps along with the their images. We resize the matrix to $14 \times 14$ using nearest neighbor down-sampling not keeping the aspect ratio. Finally, the matrix is flattened to a 196-dimensional vector and the softmax function is applied, so that $\sum{\alpha}_{i} = 1$ and $\alpha \in [0,1]$ is guaranteed like in the implementation of \citet{xu_show_2015}. An important detail is that no value is actually zero. The model is still allowed to include image aspects outside the boxes for the caption generation. In addition, when using 255 as an initial attention value, we found that we need to interpolate for each vector the pixels values following $[0,255] \rightarrow [0,1]$, because otherwise the softmax results in too much weight on individual spatial areas and leads to qualitative worse captions e.g. the model is referring to polar bear for the dog on the ground.
 
\subsection{Forcing the Spatial Attention}

The trained image captioning model has to produce what we call a \textit{box caption} for each constructed attention vector. That is, the spatial attention is derived from the bounding box like in section \ref{construct} and applied to the model in one of the following ways.

\begin{enumerate}
	
	\item[(a)] \textbf{Unlimited step-wise fixed attention.} For this experiment, we feed the spatial attention vector at each time step to the model for the whole caption generation process. The model's own predicted attention is dismissed.
	
	\item[(b)] \textbf{Limited step-wise fixed attention.} We feed the spatial attention vector for the first $i = \{3,6,9\}$ time steps which are empirically chosen. After the $t_{i}$ time step, the model is again ``free to choose'' the spatial attention depending on its state and the previous word.
	
	\item[(c)] \textbf{Step-wise additive attention.} At each time step, the spatial attention vector is added to the one predicted by the model. We introduce a factor to control the weight of the externally induced attention and divide by the according term plus 1.
	
\end{enumerate}

\section{Results}

For the test image shown in Figure~1, the model produces the caption ``a dog is laying down on the street'' with unlimited fixed attention on the dog, whereas when forcing the attention on the bicycle: ``a bicycle parked in front of a bicycle''. 

Likewise, when forcing the attention on the bike, the model is producing bike related captions ``a bicycle parked in front of a building'' for our limited step-wise configurations. In contrast to that, additive attention in this case leads to dog related captions ``a dog is sitting on the sidewalk next to a bike'' for lower weights up to two and bike related captions ``a bicycle parked in front of a building'' for weights higher than two. More examples can be found in the supplementary material.

\subsection{Quantitative Analysis: Sensitivity}

\begin{table}
	\centering
	\begin{tabular}{| l | c | c |}
		\hline
		&  \multicolumn{2}{c|}{Sensitivity} \\
		& general (diff) & method (diff) \\
		\hline
		unlimited      & 88.68 (0.55) & \textbf{52.65} (0.54)  \\
		\hline  
		limited-3      & 85.23 (0.55) & 35.20 (0.55) \\
		limited-6      & 87.90 (0.56) & 46.49 (0.55) \\
		limited-9      & \textbf{88.88} (0.55) & 51.81 (0.54) \\
		\hline  
		additive-1     & 85.51 (0.54) & 33.43 (0.51) \\
		additive-2     & 87.26 (0.54) & 41.25 (0.53) \\
		additive-3     & 85.49 (0.54) & 44.29 (0.52) \\
		\hline             
	\end{tabular}
	\caption{The degree of sensitivity as the percentage of 117,798 box captions which deviate from the control (method sens.) or self-attending (general sens.) caption in at least a single word. The differentness for the according subset is given in parentheses as WER scores.}
	\label{table:sensitive}
\end{table}

The qualitative results show that the model is capable to react to changes in its spatial attention. As a measurement for this capability, we suggest the \textit{degree of sensitivity}. Here, we quantify how often the resulting box captions deviate from the normally generated caption for an image. In the following, we call the normally generated caption a \textit{self-attending caption}, because the attention is ``freely chosen'' by the model.

As shown in Table~\ref{table:sensitive}, the model has the highest general sensitivity for the limited-9 fixed attention method where 88.88\% of the box captions differ from the self-attending caption in at least a single word. We also compute the WER scores for these subsets and see that on average the box captions are changed in every second word in comparison to the self-attending caption.

In addition, we indicate whether the model's changes in caption generation are related to specific attention forcing methods or a method unrelated phenomenon. To do so, we let the model produce a \textit{control caption} where the fixed spatial attention has been distributed uniformly over the whole image. This makes it possible to study the effect of the individual forcing methods.

The highest degree of method specific sensitivity is measured for the unlimited fixed attention method as depicted in  Table~\ref{table:sensitive}. Here, 52.65\% of the box captions differ to the control caption in at least one word. This indicates that among the presented methods, the unlimited spatial fixation is the most effective attention induction method.

\subsection{Quantitative Analysis: Controllability}

\begin{table}
	\centering
	\begin{tabular}{| l | r | r | r | r |}
		\hline
		& \multicolumn{2}{c}{Controllable} & \multicolumn{2}{c|}{and distinct}  \\
		&  k@1  & k@5   &  k@1  & k@5 \\
		\hline
		unlimited      & \textbf{28.56} & \textbf{58.17} & \textbf{9.00} & \textbf{21.39} \\
		\hline  
		limited-3      & 26.36 & 50.84 & 6.89 & 15.24  \\
		limited-6      & 27.69 & 52.75 & 8.21 & 17.86 \\
		limited-9      & 27.32 & 52.94 & 7.85 & 18.03  \\
		\hline  
		additive-1     & 25.86 & 52.89 & 6.27 & 17.23  \\
		additive-2     & 26.98 & 52.28 & 7.26 & 16.70  \\
		additive-3     & 27.35 & 53.83 & 7.33 & 18.69  \\
		\hline             
	\end{tabular}
	\caption{The degree of controllability as the percentage of box captions containing their category in relation to all 117,798 box captions. The degree for the distinct share is based on 87,033 (k@1) or 58,407 (k@5) box captions with before unmentioned categories.}
	\label{table:controlable}
\end{table}

Finally, we expect that the produced box captions are referring to objects in the bounding boxes from which the spatial attention vectors are constructed. Thus we evaluate the \textit{degree of controllability} of our forcing methods by checking that the box captions include the according box categories (k@1).\footnote{The defined metric provides a lower boundary on the performance of attention control, since we compare freely generated captions with a restricted list of classes. We leave a manual evaluation for future work.} 

Table~\ref{table:controlable} shows the highest degree of controllability for the unlimited configuration, which results in 28.56\% of the cases in a box caption that includes its according box category. As the set of COCO categories is rather restrictive, and e.g. only states ``person'' where a caption might say ``woman'' or ``man'', we also check for the five nearest neighbors in cosine distance of the model's learned word embedding space (k@5). Still, the limited-6 configuration results in the highest score with 55.58\%, in which, for example, the box caption contains the category name ``person'' or one of its neighbors (``man'', ``woman'', ``guy'' or ``girl'').

Furthermore, we compute the degree of controllability within the more interesting \textit{distinct} subset. In cases where the model  already refers to objects within the bounding box, because they include the main objects of an image, we cannot conclude whether the forcing methods have a controllable impact on the resulting captions. Thus, for the distinct subset we discard bounding boxes from the evaluation, which have categories attached that are already included in the standardly produced caption (nearest neighbors accordingly).

Table~\ref{table:controlable} shows for the distinct subset that in 9.00\% of the cases the resulting caption includes the box category, when the spatial attention is focusing on something new (not mentioned before) in the image using the unlimited fixed attention method. The unlimited configuration has also the highest degree with 21.39\%, when we also allow the five nearest neighbors to be included.
	
\section{Discussion and further work}

The results show that a caption generation model with spatial attention is controllable by the presented forcing methods. The forced model produces predictable results in up to 28.56\% of the cases. These results provide evidence that the model is inherently learning to react to changes in the spatial attention, although the learning task is a more general one. Therefore these results show that specific types of attention like spatial attention might be useful control mechanisms.

The evaluation is difficult, because we use a general purpose dataset in MSCOCO. For example, the most common category in the dataset is ``person'', which is also the most diverse one. We tried to tackle this problem by also looking for the nearest neighbors of the categories and achieved up to 21.39\% matches in the relevant subset.

Future work will include building and using cleaner and more balanced datasets for the proposed evaluation task. The model's performance is expected to improve when trained on a larger dataset like Visual Genome \cite{krishna_visual_2016}. We think that modifying the spatial attention of a standard neural image captioning model introduces an interesting new research direction for natural language generation, which will allow researchers to handle and understand the complexities of these models more easily.

\bibliography{acl2019}
\bibliographystyle{aclnatbib}

\appendix

\onecolumn

\section{Supplementary Material}

\subsection{Step-wise self-attention on the sample image}
\label{selfattention_sample}

\begin{figure*}[h]
	\centering
	\includegraphics[width=0.88\textwidth]{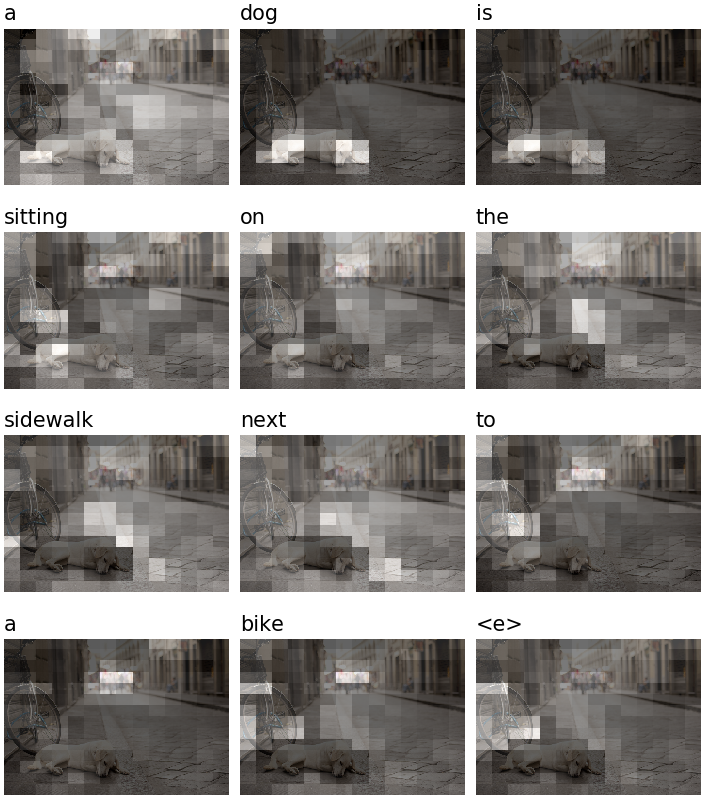}
	\caption{The step-wise alternating self-attention for the sample image with the dog and the bike. The produced caption is shown from the upper-left corner to the bottom-right corner with one word and its attention at a time-step. The models attention is shown in pixels that represent the spatial area of the visual features. The model is clearly putting higher weights on the dog, when producing the word at the second time-step. Furthermore, spatial attention is around the dog, when the resulting word is "sidewalk". When the "bike" word is produced, then also some weights are put on the persons in the background and only with the end-tag the attention is clearly on the bike.}
\end{figure*}

\clearpage

\subsection{Unlimited step-wise fixed attention on the sample image}	
\label{unlimited_sample}

\begin{figure*}[ht]
	\centering
	\subfloat[Caption: "a dog is laying down on the street"]{
		\includegraphics[width=0.4\textwidth]{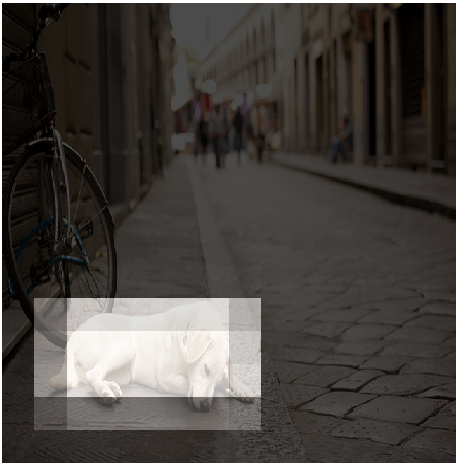}
	}
	\hfill
	\subfloat[Caption: "a bicycle parked in front of a bicycle"]{
		\includegraphics[width=0.4\textwidth]{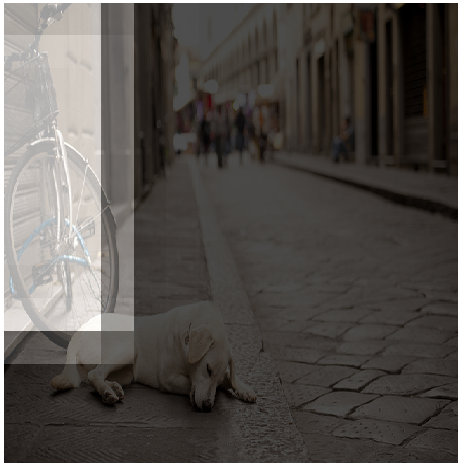}
	}
	\caption{The box captions for the sample image with fixed attention on the dog in (a) and the bicycle in (b) The model is producing captions that are clearly focusing on either the dog in (a) or the bicycle in (b).}
	
\end{figure*}

\begin{figure*}[hb]
	\centering
	\subfloat[Caption: "a giraffe standing in a pen with a person standing in the background"]{
		\includegraphics[width=0.4\textwidth]{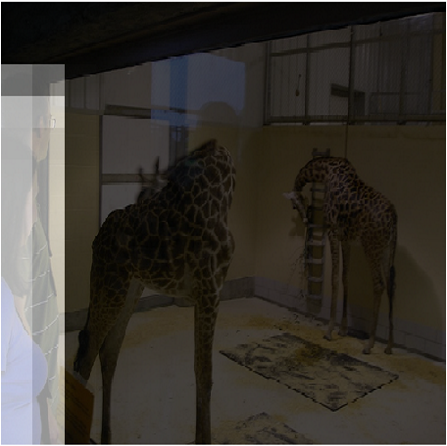}
	}
	\hfill
	\subfloat[Caption: "a giraffe standing in a pen with a giraffe"]{
		\includegraphics[width=0.4\textwidth]{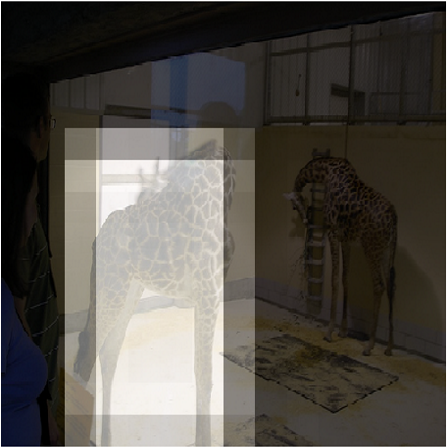}
	}
	\caption{The box captions for another sample image with person watching giraffes in a pen. The attention is either fixed on the persons in (a) on the giraffe in (b). The model is producing captions that are mentioning the persons in (a) and only the giraffes in (b).}
\end{figure*}

\clearpage

\subsection{Limited step-wise fixed attention on the sample image}
\label{semifixed_sample}

\begin{figure*}[ht]
	\centering
	\includegraphics[width=0.99\textwidth]{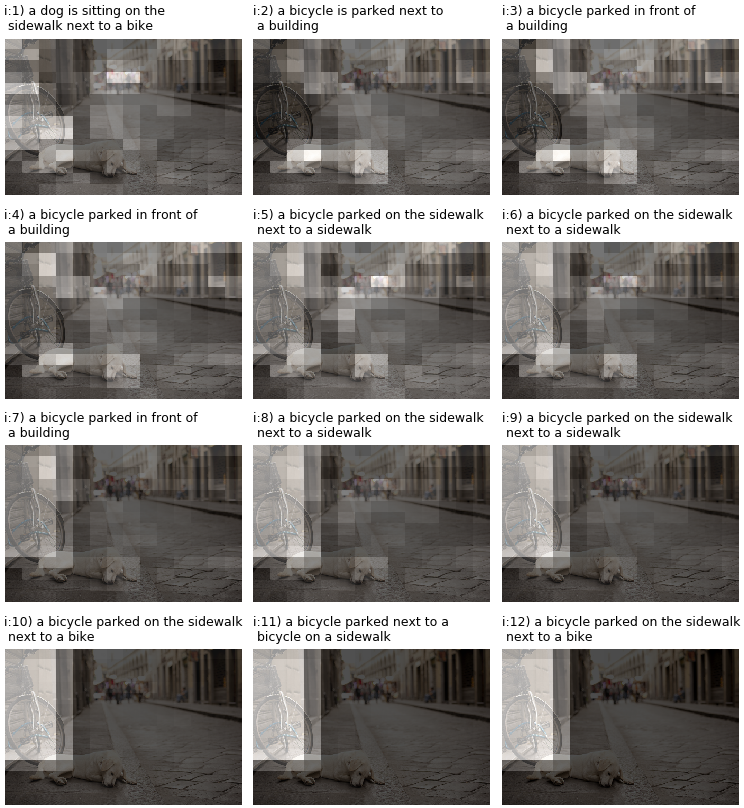}
	\caption{The limited step-wise fixed attention for the sample image with the dog and the bike. The number of fixed time-steps is increasing from the left-upper corner to the bottom-right corner from 1 to 12. The similarity with unlimited fixed attention is visible for more than nine fixed time-steps, while less than three time steps fixed are more similar to the alternating self-attention.}
\end{figure*}

\clearpage

\subsection{Additive step-wise attention on the sample image}
\label{additive_sample}

\begin{figure*}[ht]
	\centering
	\includegraphics[width=\textwidth]{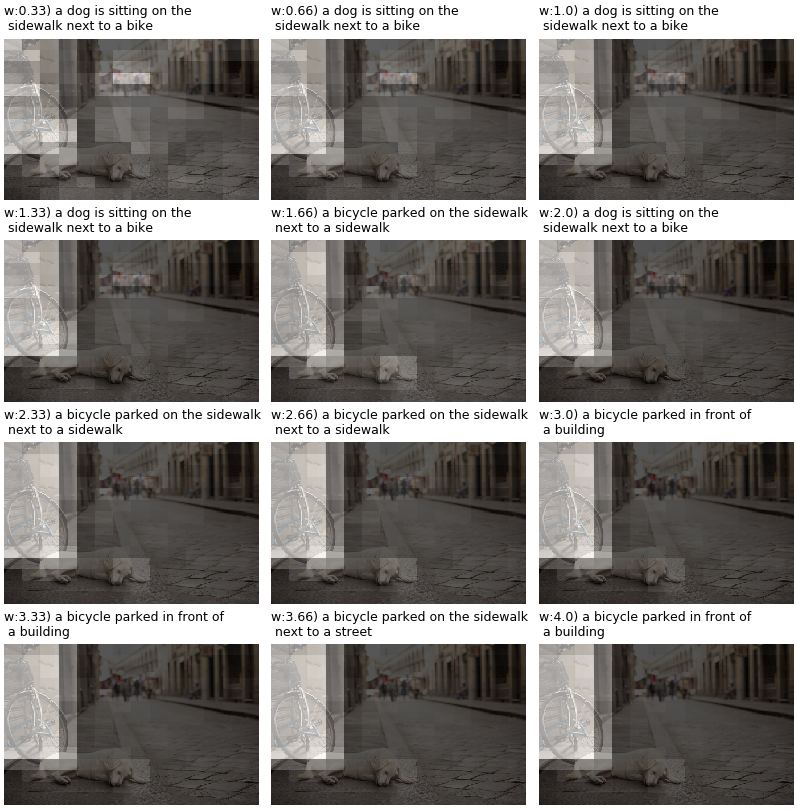}
	\caption{The step-wise additive attention for the sample image with the dog and the bike. The additive weight factor is increasing from the left-upper corner to the bottom-right corner from 0.33 to 4. The similarity with unlimited fixed attention is already visible for small factors. Nevertheless, until the weight of two most captions mention the dog first and only then the bike, while starting from a weight of two, the bicycle becomes the main object in the caption.}
\end{figure*}

\end{document}